\documentclass{article}
\usepackage{spconf,graphicx}
\usepackage{tikz}
\usetikzlibrary{positioning,spy}
\usepackage{amsmath,amsfonts,amssymb,amsthm}
\usepackage{booktabs}
\usepackage{algorithm,algorithmic}
\usepackage{bm}
\usepackage[caption=false,font=footnotesize]{subfig}
\usepackage[hidelinks,bookmarks=false]{hyperref}

\DeclareMathOperator*{\argmin}{arg\,min}
\DeclareMathOperator*{\sgn}{sgn}
\DeclareMathOperator*{\mse}{MSE}

\def\bx{\ensuremath{{\bf x}}}

\def\bg{\ensuremath{{\bf g}}}
\def\by{\ensuremath{{\bf y}}}
\def\bz{\ensuremath{{\bf z}}}
\def\bw{\ensuremath{{\bf w}}}

\def\bk{\ensuremath{{\bf k}}}
\def\bn{\ensuremath{{\bf n}}}

\def\bone{\ensuremath{{\bf 1}}}

\def\cF{\ensuremath{{\mathcal F}}}

\def\cS{\ensuremath{{\mathcal S}}}

\def\sqz{\vspace{-3pt}}
\title{An Algorithm Unrolling Approach to Deep Image Deblurring}
\name{Yuelong Li$^\dag$, Mohammad Tofighi$^\dag$, Vishal Monga$^\dag$ and Yonina C. Eldar$^\ddag$}
\vspace{-10mm}
\address{$^\dag$Department of Electrical Engineering, Pennsylvania State University, USA\\$^\dag$Department of Electrical Engineering, Technion-Israel Institute of Technology}
\begin{document}
%
\maketitle
\begin{abstract}
While neural networks have achieved vastly enhanced performance over
traditional iterative methods in many cases, they are generally empirically
designed and the underlying structures are difficult to interpret. The
algorithm unrolling approach has helped connect iterative algorithms to neural
network architectures.  However, such connections have not been made yet for
blind image deblurring. In this paper, we propose a neural network architecture
that advances this idea. We first present an iterative algorithm that may be
considered a generalization of the traditional total-variation regularization
method on the gradient domain, and subsequently unroll the half-quadratic
splitting algorithm to construct a neural network.  Our proposed deep network
achieves significant practical performance gains while enjoying
interpretability at the same time. Experimental results show that our approach
outperforms many state-of-the-art methods.
\end{abstract}
%
%
\vspace{-5mm}
\section{Introduction}\label{sec:introduction}
\vspace{-2mm}
Blind image deblurring refers to the process of recovering a sharp image from
its blurred observation. Among various deblurring problems, motion deblurring
is an important topic because camera shaking is common during photography.  Assuming a planar scene and translational camera motion, the blurring
process is typically modeled as~\cite{kundur_blind_1996}:
$\by=\bk\ast\bx+\bn$  where
$\by$ is the observed blur image, $\bx$ is the latent sharp image, $\bk$ is the
blur kernel, and $\bn$ is noise which is
often modelled as Gaussian. When $\bk$ is unknown the corresponding estimation problem is commonly called blind deconvolution.

The majority of existing blind motion deblurring methods rely on iterative
optimization.  These methods usually hinge on sparsity-inducing
regularizers, either in the gradient
domain~\cite{joshi_psf_2008,shan_high-quality_2008,cho_fast_2009,xu_two-phase_2010,krishnan_blind_2011,xu_unnatural_2013,sun_edge-based_2013,pan_$l_0$_2017}
or more general sparsifying transformation
domains~\cite{jian-feng_cai_framelet-based_2012,xiang_image_2015,pan_deblurring_2018,tofighi2018blind}.
Variants of such methods may arise indirectly from a statistical estimation
perspective, such
as~\cite{fergus_removing_2006,levin_efficient_2011,babacan_bayesian_2012}.
While these methods are typically physically interpretable, their performance
depends heavily on appropriate selection of parameters and careful design of
regularizers/priors, which are difficult to determine analytically. Furthermore,
hundreds of iterations are usually required to achieve an acceptable
performance level, and thus these algorithms can be computationally expensive.

Complementary to the aforementioned approaches, learning based methods for
determining a non-linear mapping that deblurs the image while adapting
parameter choices to an underlying training image set have been developed.
Principally important in this class are techniques that employ deep neural
networks,
including~\cite{xu_deep_2014,yan_blind_2016,chakrabarti_neural_2016,xu_motion_2018}.  Although they offer practical promises in certain scenarios such as video
deblurring and achieves substantial performance gains in some cases, these
works commonly regard neural networks as abstract function approximators.  The
structures of the networks are typically empirically determined and the actual
functionality of the neural networks is hard to interpret.

In the seminal work of Gregor {\it et al.}~\cite{gregor_learning_2010}, a novel
technique called algorithm unrolling was proposed that provides a neural network interpretation of iterative sparse coding algorithms.
Passing through the network is equivalent to executing the iterative algorithm
a finite number of times, and the trained network can be naturally interpreted
as a parameter optimized algorithm.  In blind
deblurring, Schuler {\it et al.}~\cite{schuler_learning_2016} employ neural
networks as feature extraction modules and integrate it into a trainable
deblurring system.  However, the network portions are still empirical and the
whole system remains hard to interpret.  The link between traditional
iterative algorithms and  neural networks remains largely unexplored for
the problem of blind deblurring.

In this paper, we develop a neural network approach for blind motion deblurring
in the spirit of algorithm unrolling, called Deblurring via Algorithm Unrolling
(DAU). Parameters of the algorithm are optimized by training the network and
performance gains are achieved without sacrificing interpretability. 
We experimentally verify its superior performance, both over best-known iterative algorithms and more recent neural network approaches.

\vspace{-2mm}
\section{Deblurring Via Algorithm Unrolling}\label{sec:formulation}
\vspace{-1mm}
The total-variation regularization approach in the gradient domain~\cite{perrone_clearer_2016} solves the following optimization
problem:
\vspace{-3mm}
\begin{align}
	\min_{\bk,\bg_1,\bg_2}&\frac{1}{2}\left(\left\|D_x\by-\bk\ast\bg_1\right\|_2^2+\left\|D_y\by-\bk\ast\bg_2\right\|_2^2\right)\nonumber\\
						  &+\lambda_1\|\bg_1\|_1+\lambda_2\|\bg_2\|_1+\frac{\epsilon}{2}\|\bk\|_2^2,\nonumber\\
	\text{subject to }&\bone^T\bk=1,\quad\bk\geq 0,\label{eq:gradient_optimization}
\end{align}
where $D_x\by,D_y\by$ are the partial derivates of $\by$ in
horizontal and vertical directions respectively, $\bone$ is a
vector whose entries are all ones, and $\|\cdot\|_p$ denotes the $\ell^p$
vector norm. The parameters $\lambda_1,\lambda_2,\varepsilon$ are positive
constants which balance the contributions of each term. The
$\geq$ sign acts elementwise.

In practice, $D_x\by$ and $D_y\by$ are usually computed using discrete filters,
such as the Prewitt and Sobel filters. From this viewpoint, a straightforward
generalization of~\eqref{eq:gradient_optimization} is to use more than two
filters. We formulate the generalized optimization problem as the following:
\vspace{-3mm}
\begin{align}
	\min_{\bk,{\{\bg_i\}}_i}&\sum_{i=1}^C\left(\frac{1}{2}\left\|\mathbf{f}_i\ast\by-\bk\ast\bg_i\right\|_2^2+\lambda_i\|\bg_i\|_1\right)+\frac{\epsilon}{2}{\|\bk\|}_2^2,\nonumber\\
	\text{subject to }&\|\bk\|_1=1,\quad\bk\geq 0,\label{eq:objective}
\end{align}
where ${\{\mathbf{f}_i\}}_{i=1}^C$ is a collection of $C$ filters that will be
determined subsequently through learning.
\begin{algorithm}[H]
	\renewcommand{\algorithmicrequire}{\textbf{Input:}}
	\renewcommand{\algorithmicensure}{\textbf{Output:}}
	\caption{Half-quadratic Splitting Algorithm}

	\begin{algorithmic}[1]
		\REQUIRE{Blurred image $\by$, filter banks ${\{\mathbf{f}^l_i\}}_{i,l}$}, positive constant parameters ${\{\zeta^l_i,\lambda^l_i\}}_{i,l},\varepsilon$, number of iterations $L$.
		\ENSURE{Estimated kernel $\widetilde{\bk}$, feature maps ${\{\widetilde{\bg_i}\}}_{i=1}^C$.}
		\STATE{Initialize $\bk\gets\delta$;\,$\bz_i\gets 0,i=1,\dots,C$.}
		\FOR{$l=1$ \TO $L$}
			\FOR{$i=1$ \TO $C$}
			\STATE{$\by_i^l\gets\mathbf{f}^l_i\ast\by$,}\label{step:yfilter}
				\STATE{$\bg_i^{l+1}\gets\cF^{-1}\left\{\frac{\zeta^l_i\widehat{\bk^l}^\ast\odot\widehat{\by^l_i}+\widehat{\bz_i^l}}{\zeta^l_i\left|\widehat{\bk^l}\right|^2+1}\right\}$,}\label{step:gupdate}
				\STATE{$\bz_i^{l+1}\gets\cS_{\lambda^l_i\zeta^l_i}\left\{\bg_i^{l+1}\right\}$,}\label{step:threshold}
			\ENDFOR
			\STATE{$\bk^{l+\frac{1}{3}}\gets\cF^{-1}\left\{\frac{\sum_{i=1}^C\widehat{\bz_i^{l+1}}^\ast\odot\widehat{\by^l_i}}{\sum_{i=1}^C\left|\widehat{\bz_i^{l+1}}\right|^2+\epsilon}\right\}$,}\label{step:kupdate}
			\STATE{$\bk^{l+\frac{2}{3}}\gets{\left[\bk^{l+\frac{1}{3}}\right]}_+$, $\bk^{l+1}\gets\frac{\bk^{l+\frac{2}{3}}}{\left\|\bk^{l+\frac{2}{3}}\right\|_1}$,}
		\ENDFOR
	\end{algorithmic}\label{alg:half_quadratic}
\end{algorithm}

\subsection{Efficient Minimization via Half-quadratic Splitting}\label{subsec:optimization}
A common approach to solve~\eqref{eq:gradient_optimization} and more
generally~\eqref{eq:objective} is the half-quadratic splitting
algorithm~\cite{wang_new_2008}.  The basic idea is to perform
variable-splitting and then alternating minimization on the penalty function.
To this end, we first cast~\eqref{eq:objective} into the following
approximation model:
\begin{align}
	\min_{\bk,{\{\bg_i,\bz_i\}}_i}&\sum_{i=1}^C\left(\frac{1}{2}{\left\|\mathbf{f}_i\ast\by-\bk\ast\bg_i\right\|}_2^2\right.\nonumber\\
	+&\left.\lambda_i{\|\bz_i\|}_1+\frac{1}{2\zeta_i}{\|\bg_i-\bz_i\|}_2^2\right)+\frac{\epsilon}{2}{\|\bk\|}_2^2,\nonumber\\
	\text{subject to }&\|\bk\|_1=1,\quad\bk\geq 0,\label{eq:penalty}
\end{align}
by introducing auxiliary variables ${\{\bz_i\}}_{i=1}^C$ and constant parameters $\zeta_i,i=1,\dots,C$.
We then alternately minimize over ${\{\bx_i\}}_i,{\{\bz_i\}}_i$ and
$\bk$ and iterate until convergence.

In practice, a common strategy is to alter the parameters per
iteration~\cite{wang_new_2008,xu_unnatural_2013,perrone_clearer_2016,pan_$l_0$_2017}.
In numerical analysis and optimization, this strategy is formally called
continuation method.  By adopting this strategy, we choose different parameters
${\{\zeta_i^l,\lambda_i^l\}}_{i,l}$ across the iterations $l$.  We take this
idea one step further by varying the filters ${\{\mathbf{f}_i\}}_i$ as well.
The complete algorithm is summarized in Algorithm~\ref{alg:half_quadratic}.  We
let $\widehat{\cdot}$ denote the Discrete Fourier Transform (DFT) and
$\cF^{-1}$ be the inverse DFT.\@ We define ${[x]}_+=\max\{x,0\}$,
$\delta$ is the unit impulse function,
$\cdot^\ast$ is the complex conjugation and $\odot$ is the Hadamard
 product operator. Finally, $\cS_\lambda(\cdot)$ is the soft-thresholding operator:
 $\cS_{\lambda}(x)=\sgn(x)\cdot\max\{|x|-\lambda, 0\}$. Operations
 matrices and vectors act elementwise.

After algorithm~\ref{alg:half_quadratic} converges, we obtain the
estimated feature maps ${\{\widetilde{\bg_i}\}}_i$ and the estimated
kernel $\widetilde{\bk}$.  When $\widetilde{\bk}$ approximates $\bk$,
$\widetilde{\bg_i}$ should approximate $\mathbf{f}_i\ast\bx$. Therefore,
we retrieve the image $\bx$ by solving the following optimization problem:
\begin{align}
	\widetilde{\bx}&\gets\argmin_{\bx}\frac{1}{2}\left\|\by-\widetilde{\bk}\ast\bx\right\|_2^2+\sum_{i=1}^C\frac{\eta_i}{2}\left\|\mathbf{f}_i\ast\bx-\widetilde{\bg_i}\right\|_2^2\nonumber\\
				   &=\cF^{-1}\left\{\frac{\widehat{\widetilde{\bk}}^\ast\odot\widehat{\by}+\sum_{i=1}^C\eta_i\widehat{\mathbf{f}_i}^\ast\odot\widehat{\widetilde{\bg_i}}}{\left|\widehat{\widetilde{\bk}}\right|^2+\sum_{i=1}^C\eta_i\left|\widehat{\mathbf{f}_i}\right|^2}\right\},\label{eq:image_reconstruction}
\end{align}
where $\eta_i$'s are positive constant parameters.

\begin{figure*}[t]
	\centering
	\includegraphics[width=\textwidth]{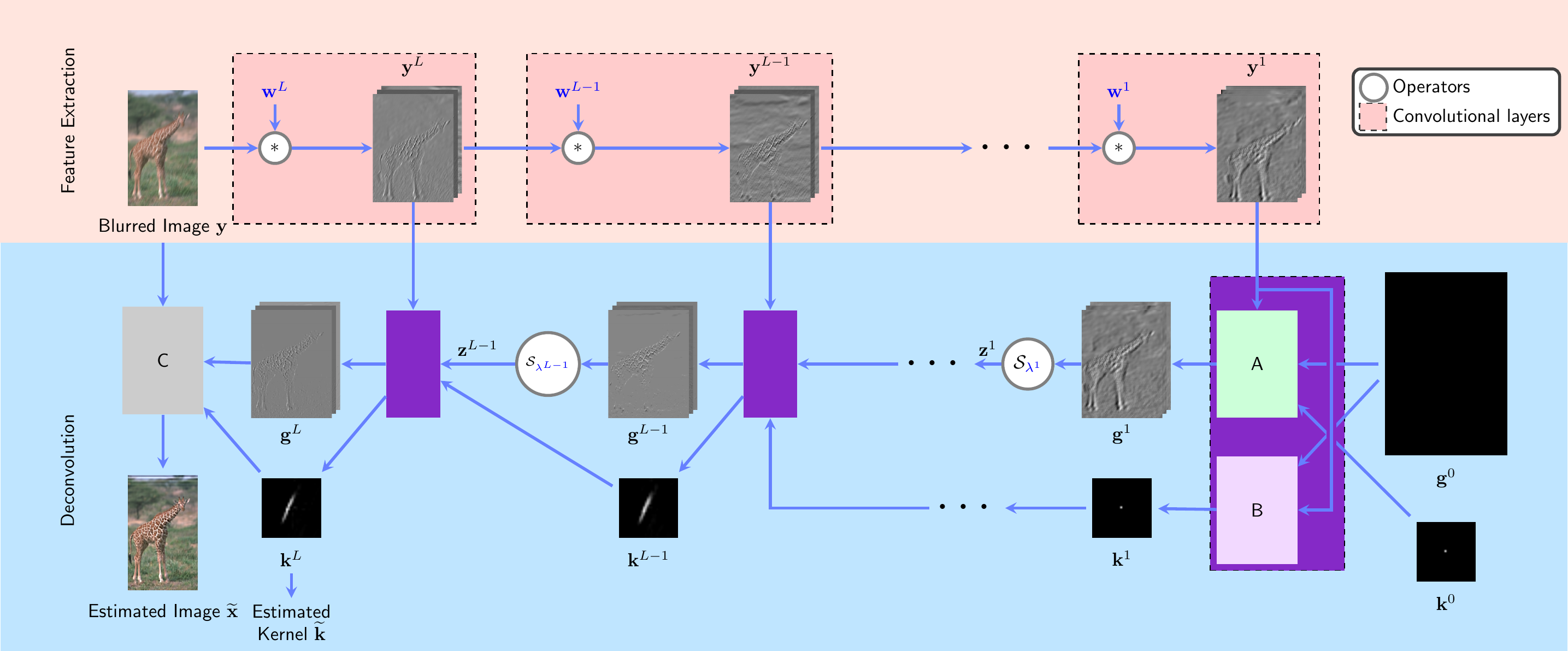}
	\caption{Structure of the deep network constructed by algorithm unrolling and cascaded filtering. Block $A$ and $B$ implements Step~\ref{step:gupdate} and Step~\ref{step:kupdate} in Algorithm~\ref{alg:half_quadratic} respectively, while block $C$ implements~\eqref{eq:image_reconstruction}. A diagram representation can be found at \url{http://signal.ee.psu.edu/diagram.pdf}. Intermediate data (hidden layers) on the trained network are also shown. It can be observed that, as $l$ increases, more details are extracted in $\bg^l$ and finer kernel coefficients are recovered. The parameters that will be learned from real datasets are colored in blue.}\label{fig:network}
	\vspace{-6mm}
\end{figure*}

\vspace{-3mm}
\subsection{Network Construction via Algorithm Unrolling}\label{subsec:unrolling}
\sqz
Each step of Algorithm~\ref{alg:half_quadratic} is in analytic form and can be
implemented using a series of basic functional operations.
Therefore, each iteration of
Algorithm~\ref{alg:half_quadratic} admits a layered representation, and
repeating it $L$ times yields an $L$-layer neural network (assuming $L$ iterations).  For
notational brevity, we concatenate the parameters in each layer and let
$\mathbf{f}^l={(\mathbf{f}_i^l)}_{i=1}^C,\zeta^l={(\zeta_i^l)}_{i=1}^C,\lambda^l={(\lambda_i^l)}_{i=1}^C$
and $\eta={(\eta_i)}_{i=1}^C$. We also concatenate $\by_i^l$'s, $\bz_i^l$'s and
$\bg_i^l$'s by letting $\by^l={(\by_i^l)}_{i=1}^C$, $\bz^l={(\bz_i^l)}_{i=1}^C$
and $\bg^l={(\bg_i^l)}_{i=1}^C$, respectively.

To handle large blur kernels, we alter the size of the
filter banks ${\{\mathbf{f}_i\}}_i$ in different layers in the following way:
\[
	\text{size of }\mathbf{f}_i^1>\text{ size of }\mathbf{f}_i^2>\text{ size of }\mathbf{f}_i^3>\dots.
\]
so that high-level representations features are captured first, and fine
details emerge in later iterations. To facilitate training, we produce
large filters by cascading small $3\times 3$ filters, following the same
principle as~\cite{simonyan_very_2015}. Formally speaking, we set
$\mathbf{f}_i^L=\bw_{i1}^L$ where ${\{\bw_{i1}^L\}}_{i=1}^C$ is a collection of
$3\times 3$ filters, and recursively obtain $\mathbf{f}_i^l$ by:
	$\mathbf{f}_i^l\gets\sum_{j=1}^C\bw_{ij}^l\ast\mathbf{f}_j^{l+1}.$
Using this representation, we obtain the network structure
in Fig.~\ref{fig:network}. The parameters ${\{\bw^l,b^l,\lambda^l\}}_{l=1}^L$ will be learned from the training data, as explained in the next Section.

\begin{figure*}[t]
	\centering
	\setcounter{subfigure}{0}
	\subfloat[Groundtruth]{%
		\begin{tikzpicture}[spy using outlines={rectangle,magnification=8,width=3.9cm,height=2.5cm}]
			\node {\includegraphics[height=0.11\textheight]{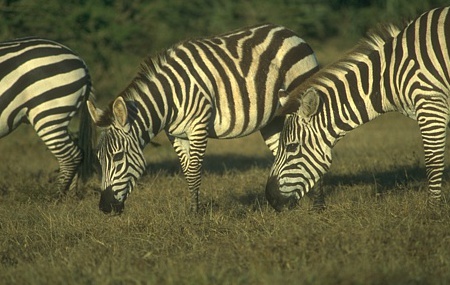}\llap{\includegraphics[height=0.03\textheight]{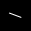}}};
			\spy[blue] on (1.5,0.7) in node[below,ultra thick] at (0,-1.5);
		\end{tikzpicture}
	}\hspace{-3mm}
	\subfloat[Perrone {\it et al.}~\cite{perrone_clearer_2016}]{%
		\begin{tikzpicture}[spy using outlines={rectangle,magnification=8,width=3.9cm,height=2.5cm}]
			\node {\includegraphics[height=0.11\textheight]{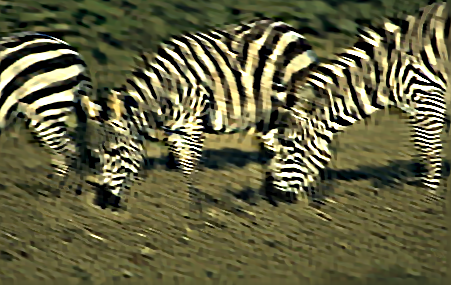}\llap{\includegraphics[height=0.03\textheight]{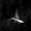}}};
			\spy[blue] on (1.5,0.7) in node[below,ultra thick] at (0,-1.5);
		\end{tikzpicture}
	}\hspace{-3mm}
	\subfloat[Nah {\it et al.}~\cite{nah_deep_2017}]{%
		\begin{tikzpicture}[spy using outlines={rectangle,magnification=8,width=3.9cm,height=2.5cm}]
			\node {\includegraphics[height=0.11\textheight]{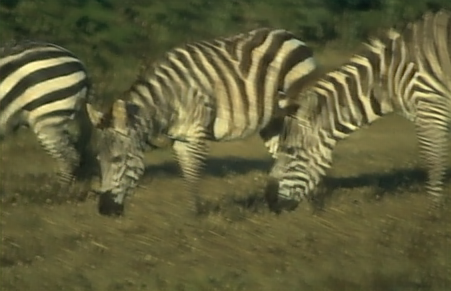}};
			\spy[blue] on (1.5,0.7) in node[below,ultra thick] at (0,-1.5);
		\end{tikzpicture}
	}\hspace{-3mm}
	\subfloat[DAU]{%
		\begin{tikzpicture}[spy using outlines={rectangle,magnification=8,width=3.9cm,height=2.5cm}]
			\node {\includegraphics[height=0.11\textheight]{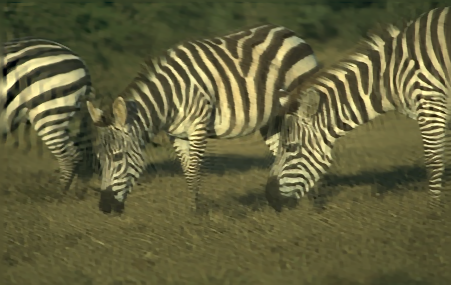}\llap{\includegraphics[height=0.03\textheight]{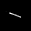}}};
			\spy[blue] on (1.5,0.7) in node[below,ultra thick] at (0,-1.5);
		\end{tikzpicture}
	}
	\caption{Qualitative comparisons on the BSDS500 dataset~\cite{martin_database_2001}. The blur kernels are placed at the right below corner. DAU recovers the kernel at higher accuracy and therefore the estimated images are more faithful to the groundtruth.}\label{fig:compare}
	\vspace{-4mm}
\end{figure*}
\sqz
\begin{figure*}[t]
	\centering
	\setcounter{subfigure}{0}
	\subfloat[Groundtruth]{%
		\begin{tikzpicture}[spy using outlines={rectangle,magnification=8,width=2.5cm,height=2cm}]
			\node {\includegraphics[height=0.11\textheight]{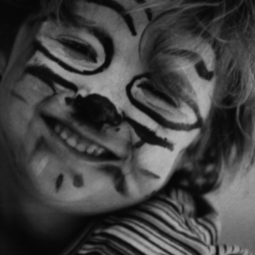}\llap{\includegraphics[height=0.03\textheight]{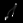}}};
			\spy[blue] on (0.3,0.2) in node[below,ultra thick] at (0,-1.4);
		\end{tikzpicture}
	}\hspace{-3mm}
	\subfloat[Perrone {\it et al.}~\cite{perrone_clearer_2016}]{%
		\begin{tikzpicture}[spy using outlines={rectangle,magnification=8,width=2.5cm,height=2cm}]
			\node {\includegraphics[height=0.11\textheight]{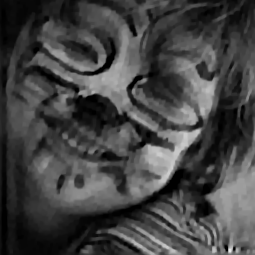}\llap{\includegraphics[height=0.03\textheight]{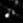}}};
			\spy[blue] on (0.3,0.2) in node[below,ultra thick] at (0,-1.4);
		\end{tikzpicture}
	}\hspace{-3mm}
	\subfloat[Nah {\it et al.}~\cite{nah_deep_2017}]{%
		\begin{tikzpicture}[spy using outlines={rectangle,magnification=8,width=2.5cm,height=2cm}]
			\node {\includegraphics[height=0.11\textheight]{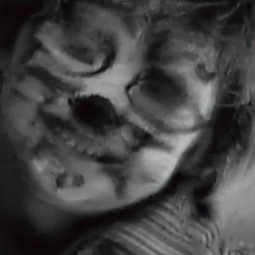}};
			\spy[blue] on (0.3,0.2) in node[below,ultra thick] at (0,-1.4);
		\end{tikzpicture}
	}\hspace{-3mm}
	\subfloat[Chakrabarti~\cite{chakrabarti_neural_2016}]{%
		\begin{tikzpicture}[spy using outlines={rectangle,magnification=8,width=2.5cm,height=2cm}]
			\node {\includegraphics[height=0.11\textheight]{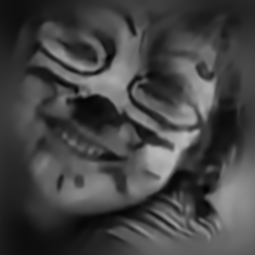}\llap{\includegraphics[height=0.03\textheight]{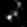}}};
			\spy[blue] on (0.3,0.2) in node[below,ultra thick] at (0,-1.4);
		\end{tikzpicture}
	}\hspace{-3mm}
	\subfloat[Xu {\it et al.}~\cite{xu_motion_2018}]{%
		\begin{tikzpicture}[spy using outlines={rectangle,magnification=8,width=2.5cm,height=2cm}]
			\node {\includegraphics[height=0.11\textheight]{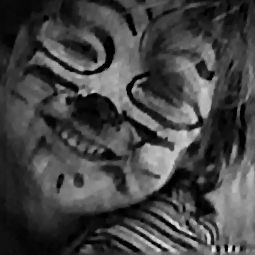}\llap{\includegraphics[height=0.03\textheight]{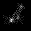}}};
			\spy[blue] on (0.3,0.2) in node[below,ultra thick] at (0,-1.4);
		\end{tikzpicture}
	}\hspace{-3mm}
	\subfloat[DAU]{%
		\begin{tikzpicture}[spy using outlines={rectangle,magnification=8,width=2.5cm,height=2cm}]
			\node {\includegraphics[height=0.11\textheight]{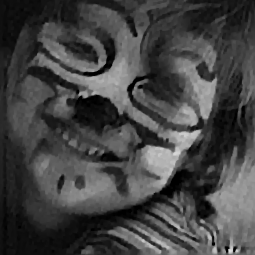}\llap{\includegraphics[height=0.03\textheight]{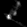}}};
			\spy[blue] on (0.3,0.2) in node[below,ultra thick] at (0,-1.4);
		\end{tikzpicture}
	}
	\caption{Qualitative comparisons on the dataset from~\cite{levin_understanding_2009}. The blur kernels are placed at the right below corner. DAU generates fewer artifacts and preserves more details than competing state of the art methods.}\label{fig:nonlinear}
	\vspace{-3mm}
\end{figure*}

\vspace{-2mm}
\section{Experiments}\label{sec:experiments}
\begin{table}
	\centering
	\caption{Quantitative comparison averaged over 200 images from the BSDS500~\cite{martin_database_2001} set and 4 linear kernels. The RMSE values are computed over kernels. Best scores are in bold.}\label{tab:scores}
	\begin{tabular}{cccc}
		\toprule
		Metrics & DAU  & \cite{perrone_clearer_2016} & \cite{nah_deep_2017}\\
		\midrule
		PSNR (dB) & $\mathbf{27.21}$ & $22.23$ & $25.23$\\
		\midrule
		ISNR (dB) & $\mathbf{4.36}$ & $2.06$ & $1.88$\\
		\midrule
		SSIM & $\mathbf{0.88}$ & $0.76$ & $0.81$\\
		\midrule
		RMSE ($\times10^{-3}$) & $\mathbf{2.21}$ & $5.21$ & $-$\\
		\bottomrule
	\end{tabular}
\end{table}
\begin{table}
	\centering
	\caption{Quantitative comparison on nonlinear motion (average over 4 images and 8 kernels from~\cite{levin_understanding_2009}). The RMSE values are computed over kernels. Best scores are in bold.}\label{tab:nonlinear}
	\begin{tabular}{cccccc}
		\toprule
		& DAU & \cite{perrone_clearer_2016} & \cite{nah_deep_2017} & \cite{chakrabarti_neural_2016} & \cite{xu_motion_2018}\\
		\midrule
		PSNR (dB) & $\mathbf{27.15}$ & $26.79$ & $24.51$ & $23.18$ & $26.75$\\
		\midrule
		ISNR (dB) & $\mathbf{3.79}$ & $3.63$ & $1.35$ & $0.02$ & $3.59$\\
		\midrule
		SSIM & $0.88$ & $\mathbf{0.89}$ & $0.81$ & $0.81$ & $\mathbf{0.89}$\\
		\midrule
		RMSE ($\times 10^{-3}$) & $3.87$ & $\mathbf{3.83}$ & $-$ & $-$ & $3.98$\\
		\bottomrule
	\end{tabular}
\end{table}
\vspace{-2mm}
\noindent \textbf{Training}: We use the training and validation portions from the Berkeley Segmentation Dataset
(BSDS500)~\cite{martin_database_2001} as training images.
The linear motion kernels are generated by uniformly sampling $16$ angles in
$[0,\pi]$ and $16$ lengths in $[5,20]$.
The images are convolved with each
kernel and white Gaussian noise with standard deviation $0.01$ (suppose the image intensity is in $[0,1]$) is added.
For each blurred image
$\by_t^\mathrm{train}(t=1,\dots,T)$, we let the corresponding sharp image and
kernel be $\bx_t^\mathrm{train}$ and $\bk_t^\mathrm{train}$, respectively.
We re-parametrize $\lambda_i^l$ in
step~\ref{step:threshold} of Algorithm~\ref{alg:half_quadratic} by letting
$b_i^l=\lambda_i^l\zeta_i^l$ and let $b^l={(b_i^l)}_{i=1}^C,l=1,\dots,L$.  The
network outputs $\widetilde{\bx}_t,\widetilde{\bk}_t$ corresponding to
$\by^\mathrm{train}_t$ depend on the network parameters $\bw^l$, $b^l,\lambda^l$,
$l=1,2,\dots,L$, and $\widetilde{\bx}_t$ further depends on $\eta$. We train the
network to determine those parameters by minimizing:
\sqz
\begin{align*}
	\min_{{\{\bw^l,b^l,\lambda^l\}}_{l=1}^L,\eta}&\sum_{t=1}^T\mse\left(\bx_t^{\mathrm{train}}-\widetilde{\bx}_t\left({\{\bw^l,b^l,\lambda^l\}}_{l=1}^L,\eta\right)\right)\\
	+&\kappa\mse\left(\bk_t^{\mathrm{train}}-\widetilde{\bk}_t\left({\left\{\bw^l,b^l,\lambda^l\right\}}_{l=1}^L\right)\right),\\
	\text{subject to }&b_i^l\geq0,\,\lambda_i^l\geq 0,\quad l=1,\dots,L,i=1,\dots,C,
\end{align*}
where $\kappa>0$ is a constant parameter which is fixed to $10^5$ and $\mse$ is the Mean Squared Error. We
choose $L=10$ and $C=16$ by cross-validation.  The minimization is performed
by stochastic gradient descent, followed by a gradient projection step to enforce the non-negative constraints. We use the Adam~\cite{kingma_adam:_2015} solver for faster training.
The learning rate is set to $1\times 10^{-3}$ initially and decayed
by a factor of $0.9$ per epoch.  We terminate training after $20$
epochs. The parameters ${\{\lambda_i^l\}}_{i,l}$ are initialized to
zeros, ${\{b_i^l\}}_{i,l}$ to $1$, and ${\{\eta_i\}}_i$
to $20$, respectively. The weights are initialized according
to~\cite{glorot_understanding_2010}.

\noindent \textbf{Evaluation}:
We use $200$ images from the test portion from the BSDS500
dataset~\cite{martin_database_2001} as test images. We randomly choose angles
from $[0,\pi]$ and lengths from $[5,20]$ to generate $4$ test kernels.  We
compare with state-of-the art algorithms, Perrone {\it et
al.}~\cite{perrone_clearer_2016} and Nah {\it et
al.}~\cite{nah_deep_2017}, which are representatives of iterative algorithms and deep-learning approaches. We assess the performance using four commonly
used evaluation metrics: Peak Signal-to-Noise-Ratio (PSNR), Improvement in
Signal-to-Noise-Ratio (ISNR), Structural Similarity Index
(SSIM)~\cite{wang_image_2004}, and Root-Mean-Square Error (RMSE) between the
estimated kernel and the groundtruth kernel.  The average scores are in
Table~\ref{tab:scores}. Clearly, DAU outperforms state-of-the art
algorithms by a significant margin.  

\vspace{-2mm}
Fig.~\ref{fig:compare} shows example images and kernels for a qualitative
comparison. Although Perrone {\it et al.}'s method can roughly infer the
directions of the blur kernels, the recovered coefficients are unsatisfactory.
As a result, the recovered image contains clearly visible artifacts. Nah {\it
et al.}'s method effectively removes most of the blurs, but blurring artifacts
still remain locally and the details are not faithfully preserved.  In
contrast, the kernel recovered by DAU is closer to the ground truth
 and hence leads to a more accurate estimated image.

Additionally, we compare the performance of various methods on deblurring under
non-linear motion kernels, which is a more realistic scenario as discussed
in~\cite{levin_understanding_2009}. We collect training kernels by
interpolating the paths provided by~\cite{kohler_recording_2012} and created by
ourselves in the same manner. We further augment these kernels by scaling and
rotations. We use the standard image set from~\cite{levin_understanding_2009}
(comprising 4 images and 8 kernels) as the test set. The average scores are
presented in Table~\ref{tab:nonlinear}. Again DAU outperforms state-of-the-art
methods. A visual example is shown in Fig.~\ref{fig:nonlinear},


\vspace{-3mm}
\section{Conclusion}\label{sec:conclusion}
\vspace{-2mm}
We propose a neural network deblurring architecture built by unrolling an iterative
algorithm. We show how a generalized TV-regularized algorithm can be recast
into a neural network, and train it to optimize the parameters. Unlike
most existing deblurring networks, our work has the benefit of
interpretability, while exhibiting performance benefits that are shared with modern deep-nets and exceed state of the art performance.

\bibliographystyle{IEEEbib}
\ninept
\bibliography{Deconvolution,DeepLearning}

\begin{thebibliography}{10}

\bibitem{kundur_blind_1996}
D.~Kundur and D.~Hatzinakos,
\newblock ``Blind image deconvolution,''
\newblock {\em IEEE Signal Process. Mag.}, vol. 13, no. 3, pp. 43--64, May
  1996.

\bibitem{joshi_psf_2008}
N.~Joshi, R.~Szeliski, and D.~J. Kriegman,
\newblock ``{PSF} estimation using sharp edge prediction,''
\newblock in {\em Proc. {IEEE} {Conf.} {CVPR}}, June 2008.

\bibitem{shan_high-quality_2008}
Qi~Shan, Jiaya Jia, and Aseem Agarwala,
\newblock ``High-quality {Motion} {Deblurring} from a {Single} {Image},''
\newblock in {\em Proc. {ACM} {SIGGRAPH}}, 2008.

\bibitem{cho_fast_2009}
Sunghyun Cho and Seungyong Lee,
\newblock ``Fast {Motion} {Deblurring},''
\newblock in {\em Proc. {ACM} {SIGGRAPH} {Asia}}, 2009.

\bibitem{xu_two-phase_2010}
Li~Xu and Jiaya Jia,
\newblock ``Two-phase kernel estimation for robust motion deblurring,''
\newblock in {\em Proc. ECCV}, 2010.

\bibitem{krishnan_blind_2011}
Dilip Krishnan, Terence Tay, and Rob Fergus,
\newblock ``Blind deconvolution using a normalized sparsity measure,''
\newblock in {\em Proc. IEEE Conf. CVPR}, 2011.

\bibitem{xu_unnatural_2013}
L.~Xu, S.~Zheng, and J.~Jia,
\newblock ``Unnatural {L}0 {Sparse} {Representation} for {Natural} {Image}
  {Deblurring},''
\newblock in {\em Proc. IEEE Conf. CVPR}, June 2013.

\bibitem{sun_edge-based_2013}
L.~Sun, S.~Cho, J.~Wang, and J.~Hays,
\newblock ``Edge-based blur kernel estimation using patch priors,''
\newblock in {\em Proc. IEEE ICCP}, Apr. 2013.

\bibitem{pan_$l_0$_2017}
J.~Pan, Z.~Hu, Z.~Su, and M.~H. Yang,
\newblock ``\${L}\_0\$ -{Regularized} {Intensity} and {Gradient} {Prior} for
  {Deblurring} {Text} {Images} and {Beyond},''
\newblock {\em IEEE Trans. Pattern Anal. Mach. Intell.}, vol. 39, no. 2, pp.
  342--355, Feb. 2017.

\bibitem{jian-feng_cai_framelet-based_2012}
{Jian-Feng Cai}, {Hui Ji}, {Chaoqiang Liu}, and {Zuowei Shen},
\newblock ``Framelet-{Based} {Blind} {Motion} {Deblurring} {From} a {Single}
  {Image},''
\newblock {\em IEEE Trans. Image Process.}, vol. 21, no. 2, pp. 562--572, Feb.
  2012.

\bibitem{xiang_image_2015}
Shiming Xiang, Gaofeng Meng, Ying Wang, Chunhong Pan, and Changshui Zhang,
\newblock ``Image {Deblurring} with {Coupled} {Dictionary} {Learning},''
\newblock {\em Int. J. Comput. Vis.}, vol. 114, no. 2-3, pp. 248--271, Sept.
  2015.

\bibitem{pan_deblurring_2018}
J.~Pan, D.~Sun, H.~Pfister, and M.~H. Yang,
\newblock ``Deblurring {Images} via {Dark} {Channel} {Prior},''
\newblock {\em IEEE Trans. Pattern Anal. Mach. Intell.}, vol. PP, no. 99, pp.
  1--1, 2018.

\bibitem{tofighi2018blind}
M.~Tofighi, Y.~Li, and V.~Monga,
\newblock ``Blind image deblurring using row--column sparse representations,''
\newblock {\em IEEE Signal Processing Letters}, vol. 25, no. 2, pp. 273--277,
  2018.

\bibitem{fergus_removing_2006}
Rob Fergus, Barun Singh, Aaron Hertzmann, Sam~T. Roweis, and William~T.
  Freeman,
\newblock ``Removing {Camera} {Shake} from a {Single} {Photograph},''
\newblock in {\em Proc. {ACM} {SIGGRAPH}}, New York, NY, USA, 2006.

\bibitem{levin_efficient_2011}
A.~Levin, Y.~Weiss, F.~Durand, and W.~T. Freeman,
\newblock ``Efficient marginal likelihood optimization in blind
  deconvolution,''
\newblock in {\em Proc. {IEEE} {Conf.} {CVPR}}, June 2011.

\bibitem{babacan_bayesian_2012}
S.~Derin Babacan, Rafael Molina, Minh~N. Do, and Aggelos~K. Katsaggelos,
\newblock ``Bayesian {Blind} {Deconvolution} with {General} {Sparse} {Image}
  {Priors},''
\newblock in {\em Proc. {ECCV}}, Oct. 2012.

\bibitem{xu_deep_2014}
Li~Xu, Jimmy~SJ Ren, Ce~Liu, and Jiaya Jia,
\newblock ``Deep convolutional neural network for image deconvolution,''
\newblock in {\em Proc. NIPS}, 2014.

\bibitem{yan_blind_2016}
R.~Yan and L.~Shao,
\newblock ``Blind {Image} {Blur} {Estimation} via {Deep} {Learning},''
\newblock {\em IEEE Trans. Image Process.}, vol. 25, no. 4, pp. 1910--1921,
  Apr. 2016.

\bibitem{chakrabarti_neural_2016}
Ayan Chakrabarti,
\newblock ``A {Neural} {Approach} to {Blind} {Motion} {Deblurring},''
\newblock in {\em Proc. {ECCV}}, Oct. 2016.

\bibitem{xu_motion_2018}
X.~Xu, J.~Pan, Y.~J. Zhang, and M.~H. Yang,
\newblock ``Motion {Blur} {Kernel} {Estimation} via {Deep} {Learning},''
\newblock {\em IEEE Trans. Image Process.}, vol. 27, no. 1, pp. 194--205, Jan.
  2018.

\bibitem{gregor_learning_2010}
Karol Gregor and Yann LeCun,
\newblock ``Learning fast approximations of sparse coding,''
\newblock in {\em Proc. ICML}, 2010.

\bibitem{schuler_learning_2016}
Christian~J. Schuler, Michael Hirsch, Stefan Harmeling, and Bernhard Scholkopf,
\newblock ``Learning to {Deblur},''
\newblock {\em IEEE Trans. Pattern Anal. Mach. Intell.}, vol. 38, no. 7, pp.
  1439--1451, July 2016.

\bibitem{perrone_clearer_2016}
Daniele Perrone and Paolo Favaro,
\newblock ``A {Clearer} {Picture} of {Total} {Variation} {Blind}
  {Deconvolution},''
\newblock {\em IEEE Trans. Pattern Anal. Mach. Intell.}, vol. 38, no. 6, pp.
  1041--1055, June 2016.

\bibitem{wang_new_2008}
Y.~Wang, J.~Yang, W.~Yin, and Y.~Zhang,
\newblock ``A {New} {Alternating} {Minimization} {Algorithm} for {Total}
  {Variation} {Image} {Reconstruction},''
\newblock {\em SIAM J. Imaging Sci.}, vol. 1, no. 3, pp. 248--272, Jan. 2008.

\bibitem{simonyan_very_2015}
Karen Simonyan and Andrew Zisserman,
\newblock ``Very deep convolutional networks for large-scale image
  recognition,''
\newblock in {\em Proc. ICLR}, 2015.

\bibitem{nah_deep_2017}
Seungjun Nah, Tae~Hyun Kim, and Kyoung~Mu Lee,
\newblock ``Deep multi-scale convolutional neural network for dynamic scene
  deblurring,''
\newblock in {\em Proc. IEEE Conf. {CVPR}}, 2017, vol.~1, p.~3.

\bibitem{martin_database_2001}
D.~Martin, C.~Fowlkes, D.~Tal, and J.~Malik,
\newblock ``A database of human segmented natural images and its application to
  evaluating segmentation algorithms and measuring ecological statistics,''
\newblock in {\em Proc. IEEE ICCV}, July 2001.

\bibitem{levin_understanding_2009}
Anat Levin, Yair Weiss, Fredo Durand, and William~T. Freeman,
\newblock ``Understanding and evaluating blind deconvolution algorithms,''
\newblock in {\em Computer {Vision} and {Pattern} {Recognition}, 2009. {CVPR}
  2009. {IEEE} {Conference} on}. 2009, pp. 1964--1971, IEEE.

\bibitem{kingma_adam:_2015}
Diederik~P. Kingma and Jimmy Ba,
\newblock ``Adam: {A} method for stochastic optimization,''
\newblock in {\em Proc. ICLR}, 2015.

\bibitem{glorot_understanding_2010}
Xavier Glorot and Yoshua Bengio,
\newblock ``Understanding the difficulty of training deep feedforward neural
  networks,''
\newblock in {\em Proc. ICAIS}, Mar. 2010.

\bibitem{wang_image_2004}
Zhou Wang, A.C. Bovik, H.R. Sheikh, and E.P. Simoncelli,
\newblock ``Image quality assessment: from error visibility to structural
  similarity,''
\newblock {\em IEEE Trans. Image Process.}, vol. 13, no. 4, pp. 600--612, Apr.
  2004.

\bibitem{kohler_recording_2012}
Rolf Köhler, Michael Hirsch, Betty Mohler, Bernhard Schölkopf, and Stefan
  Harmeling,
\newblock ``Recording and playback of camera shake: {Benchmarking} blind
  deconvolution with a real-world database,''
\newblock in {\em European {Conference} on {Computer} {Vision}}. 2012, pp.
  27--40, Springer.

\end{thebibliography}

\end{document}